\documentclass[conference]{IEEEtran}
\IEEEoverridecommandlockouts
\usepackage{cite}
\usepackage{amsmath,amssymb,amsfonts}
\usepackage{algorithmic}
\usepackage{graphicx}
\usepackage{textcomp}
\usepackage{xcolor}
\usepackage[scientific-notation=true]{siunitx}
\def\BibTeX{{\rm B\kern-.05em{\sc i\kern-.025em b}\kern-.08em
    T\kern-.1667em\lower.7ex\hbox{E}\kern-.125emX}}
\begin{document}

\title{Novel Reinforcement Learning Algorithm for Suppressing Synchronization in Closed Loop Deep Brain Stimulators}

\author{\IEEEauthorblockN{Harsh Agarwal}
\IEEEauthorblockA{\textit{Department of Electrical and Computer Engineering} \\
\textit{Indian Institute of Technology}\\
Jodhpur, India\\
agarwal.10@iitj.ac.in}
\and
\IEEEauthorblockN{Heena Rathore}
\IEEEauthorblockA{\textit{Department of Computer Science} \\
\textit{Texas State University}\\
San Marcos, USA \\
heena.rathore@ieee.org}
}

\maketitle

\begin{abstract}
Parkinson's disease is marked by altered and increased firing characteristics of pathological oscillations in the brain. In other words, it causes abnormal synchronous oscillations and suppression during neurological processing. In order to examine and regulate the synchronization and pathological oscillations in motor circuits, deep brain stimulators (DBS) are used. Although machine learning methods have been applied for the investigation of suppression, these models require large amounts of training data and computational power, both of which pose challenges to resource-constrained DBS. This research proposes a novel reinforcement learning (RL) framework for suppressing the synchronization in neuronal activity during episodes of neurological disorders with less power consumption. The proposed RL algorithm comprises an ensemble of a temporal representation of stimuli and a twin-delayed deep deterministic (TD3) policy gradient algorithm. We quantify the stability of the proposed framework to noise and reduced synchrony using RL for three pathological signaling regimes: regular, chaotic, and bursting, and further eliminate the undesirable oscillations. Furthermore, metrics such as evaluation rewards, energy supplied to the ensemble, and the mean point of convergence were used and compared to other RL algorithms, specifically the Advantage actor critic (A2C), the Actor critic with Kronecker-featured trust region (ACKTR), and the Proximal policy optimization (PPO).
\end{abstract}


\section{Introduction}
The basal ganglia region in the brain is responsible for motor control, eye movement, and other cognitive and emotional functionality~\cite{bg}. In Parkinson's disease, the basal ganglia region is altered to produce increased firing rates and oscillations in the brain~\cite{pd}. Deep brain stimulators (DBS) are used to treat these altered oscillations when medications are clinically ineffective and patients show refractory symptoms~\cite{pi}. To-date, DBS have been implanted on over 160,000 people globally to treat a variety of neurological and non-neurological disorders~\cite{rat}.  

The circuitry of present open loop DBS devices does not include any feedback algorithms; instead, clinicians simply alter the electrode currents in response to clinical observations~\cite{krylov2020}. According to preliminary clinical research, closed-loop DBS for Parkinson's disease has the potential to provide patients with better symptom and side effect control while using less power than open-loop DBS~\cite{fle}. On the other hand, computational models can help to understand how the system behaves in response to stimulation, provide insights into potential mechanisms of action, and offer a platform for testing closed-loop control strategies~\cite{Lu2020}. Furthermore, the advancement of reinforcement learning (RL) in recent years has offered a feedback-based control system that is extensively used for analysis and learning based on the feedback given by the environment~\cite{neuman}.
 
 Besides the synthetic physical modeling~\cite{12}, several RL models~\cite{Q.Gao2020}-\cite{krylov2020} have been utilized as a data-driven model to suppress the synchronized oscillating neuronal activity for DBS. However, traditional RL models in DBS have less suppression coefficient and a higher mean point of convergence of oscillations. Furthermore, actor-critic based RL models suffer from bias and variance trade-off in the target values~\cite{issue}. To overcome these issues, we specifically model the improved version of actor critic reinforcement learning model for closed loop DBS. The proposed model can bring about a significant decrease in the synchronization of neurons and provide robust data-driven control, thereby alleviating the symptoms of the neurological disorder.

The organization of the paper is as follows: Section~\ref{1k2} gives details on the building blocks of the proposed work and describes the related work in the area of modeling DBS. In Section~\ref{14}, we present our proposed algorithm. Section~\ref{15} illustrates the performance analysis of the proposed work, followed by the conclusions and future work in Section~\ref{co}.

\section{Related Work}
This section presents the recent work which uses machine learning techniques for suppressing the synchronization using DBS. Gao et al.~\cite{Q.Gao2020} proposed a model for the basal ganglia region that considered the region as a Markov decision process whose state and action spaces are the pre-processed signals from the basal ganglia region and the stimulated pulses, respectively. Convolutional neural networks were used to extract features from the pre-processed time series. Their RL-based DBS control approach reduces Parkinson's disease symptoms with low stimulation frequencies and optimal patterns. 

Authors in~\cite{Lu2020}, proposed Cerebellar Model Articulation Controller (CMAC)-based actor-critic RL framework for closed-loop DBS. The CMAC layers based on the neuro-physiological understanding of the cerebellum are incorporated in both the actor and critic, which results in fast learning as it does not involve the use of activation functions. Temporal difference method was used as the learning algorithm along with recursive least squares method for the critic so as to enhance the computing time and efficiency. 

Liu et al.~\cite{liu2020} modified the basal ganglia model proposed by Lu et al.~\cite{lu2017} which is based on Rubin and Terman's model, by randomizing the different inhibitory and excitatory connections between neurons of different regions, namely GPi, STN, TH and PY. Their work also incorporated the synaptic delay effect into their model. These models do not take into account the continuous action space and are sensitive and unstable to changes with poor convergence. Sarikhani et al.~\cite{neuroweaver} introduced an open-source platform that eases the process of designing and deploying closed-loop neuromodulation systems. Its pipeline consists of three main blocks comprising of a simulation environment for RL algorithms, libraries to support the development of models, and a cross-domain accelerator for deploying these neuromodulation systems. Rosenblum~\cite{chaos_rosenblum} proposed to modulate the synchrony of a globally coupled environment by utilizing action pulses with minimal intervention. The goal is to trigger only at the most sensitive phase, found by adaptive feedback control. 

The authors in~\cite{mann} used machine learning approaches like gradient boosted tree learning along with feature importance analysis to predict neuronal oscillations. Furthermore, the authors in~\cite{bout} used the fMRI patterns of 39 parkinson disease patients with a prior clinically adjusted DBS (with 88\% accuracy), to create machine learning model that predicts optimal vs. non-optimal settings. Although these models are used for accurate prediction, the studies lack the work to suppress the oscillations.  
\label{1k2}
The algorithm proposed in this paper aims to address the following shortcomings in state-of-the-art algorithms:
\begin{itemize}
    \item High variance in target values, which results in low values for suppression coefficients, when algorithms are run for the same number of time steps.
    \item Avoid overestimated bias to stabilize the synchronization and mean point of convergence to generalize the model well.
    \item Error build from each iteration which leads to unstable rewards. The proposed model adds noise to the target action to exploit the Q-function errors by smoothing out Q-values along with changes in the action.
    \item Results are shown for algorithms on simulators which do not capture the characteristics of closed loop DBS.
\end{itemize}
Specifically, to address the last issue, this paper uses a gym environment~\cite{brockman2016} comprising of different discrete neuro-scientific models in which neurons are governed by their specific differential equations~\cite{krylov2020}. Their work primarily focuses on two neuronal models: Bonhoeffer–van der Pol oscillators~\cite{Bonhoeffer1948} and the Bursting Hindmarsh–Rose neuronal model~\cite{hindmarsh1984} which qualitatively reproduce various patterns of collective neuronal activity, namely regular, chaotic, and bursting signaling regimes.

\section{Methods}
\label{14}

To mitigate the oscillations and prevent the overestimation of Q-values, we use a pair of actors and critics to form a twin-delayed deep deterministic (TD3) policy gradient algorithm~\cite{lilicrap2015}. The proposed RL agent was trained by employing TD3 Policy Gradient~\cite{fujimoto2018}, an off-policy RL algorithm built over DDPG~\cite{lilicrap2015} and DQN~\cite{mnih2015}. The model consists of two major components: an actor and a critic that interact with each other. Each input signal from the environment is transformed into its temporal representation and then fed to both the actor and the critic. The actor performs an action, which is fed to the critic for feedback. Based on the critic's feedback, the actor responds with an action, to which random Gaussian noise is added and observations are stored. Following are the blocks of the proposed RL algorithm:
\subsubsection{Temporal representation of stimulus}
Pavlov~\cite{pavlov} conducted an experiment wherein he trained a dog by striking a bell. Each strike of the bell was accompanied by the administration of food. Here, the sound of the bell acts as the stimulus, and food is analogous to the reward. 
This led to the hypothesis that Pavlovian learning is related to the unpredictability of reward. Furthermore, Suri and Schultz~\cite{suri1999} asserted that prolonged intervals in rewards are difficult to learn in comparison to shorter ones. Accordingly, the temporal representation of a stimulus that contains longer phasic components and shorter tonic ones would result in sustained neuronal activity. This representation helps in emulating dopamine suppression when an anticipated reward is omitted~\cite{suri2002} thereby enhancing learning. We implement this representation by breaking the neuronal stimulus into various parts, with each part having an equal but lesser magnitude than the original signal. There exists a time difference between the onset of consecutive parts. We propose the following equation to denote the temporal representation for a particular signal $(X(t))$:
\begin{equation}
 X(t) = \sum_{n=1}^{3} 0.33*(Z_n(t)) \\
 \begin{aligned}
\end{aligned}
\end{equation}
where $Z_n(t)$ denotes the time differed components of $X(t)$.
\subsubsection{Target Policy Smoothing Regularization}
Target policy smoothing, which imitates the learning updates of SARSA (Sutton and Barto~\cite{sutton1999}) is one of the enhancements introduced in the TD3 Algorithm. When updating the critic, deterministic policy techniques have a higher variance in target values. Target policy smoothing, a regularisation technique, is constructed on the presumption that alike actions correspond to alike values which is fulfilled by infusing the target with a modest amount of noise ($\epsilon$) and averaging over mini-batches as represented in (Equation 2)~\cite{fujimoto2018}. As observed by Krylov et al.~\cite{krylov2020} noisy actions have a greater impact on the agent's effectiveness, and adding noise of the similar order of magnitude as the action to the agent improves the suppression coefficient. Accordingly, statistically independent gaussian noise with a standard deviation of 0.1 and a zero mean value was appended to the model. However, as explained by Uhlenbeck and Ornstein~\cite{ornstein1930} this did not lead to significant improvements. This resembles the neuronal noise gathered due to the stochastic nature of cells that process information. This noise is bound to increase with inter-network interactivity and nonlinear computations. In spike-generating neurons, the signals that are below a particular threshold do not affect the output. Neuronal noise helps in increasing the proximity between the sub-threshold values, which introduces a smoothed non-linearity. This optimizes neural-network function by promoting spike initiation, and the developed neural networks will be robust in nature. Such networks would be exploratory and showcase increased adaptability to the evolving needs of an ever-changing environment~\cite{faisal2008}. Hence, as shown in Eq. (2), $\epsilon$ is added to the target policy, which denotes Gaussian noise.
\begin{equation}
    \begin{aligned}
    y &= R + \gamma Q\textsubscript{$\theta$'}(S',\pi\textsubscript{$\phi$'}(S') + \epsilon)
    \end{aligned}
\end{equation}
where $\epsilon$ $\sim$ clip($\mathcal{N}$(0, $\sigma$), -c, c), $R$ is the received reward, $S'$ is the new state, $\pi$\textsubscript{$\phi$} is the policy with parameters $\phi$, $\gamma$ denotes the discount factor and Q\textsubscript{$\theta$'} represents the critic network value function. 
In our work, the reward function is dependent on the mean field and the action of the RL agent. It is inversely dependent on the magnitude of action taken and the square of the difference in mean fields. The proposed reward function penalizes the onset of a large action and favors convergence to the mean field of previous states.
\begin{equation}
    \begin{aligned}
    R(t) &= \frac{100}{(X(t) -[X\textsubscript{state}]\textsubscript{t}])^2 +10} + \frac{3}{A + 10}
    \end{aligned}
\end{equation}

\subsubsection{Twin Critic Network}
This improvement was inspired by Double Q-Learning~\cite{vanhasselt2016} that comprised predicting the current Q-value using a separate target value function to reduce bias. Two different value estimates are preserved, wherein each of these is employed to keep the other updated. This ensures unbiased estimations of the actions chosen using the opposing value estimate. Nevertheless, for actor-critic methods, this methodology is non-functional as the policy and target networks are modified so sluggishly that they appear to be identical, which again leads to overestimation. Instead, clipped double Q-Learning (Eq. 2) is implemented with a pair of critics and an actor to address this issue~\cite{fujimoto2018}. The two critics are not absolutely independent due to the replay buffer and the use of opposite critics for updates. At times it may happen that second critic might estimate more than first critic~\cite{fujimoto2018}. 
\begin{equation}
    \begin{aligned}
    Q\textsubscript{$\theta$\textsubscript{2}}(S,\pi\textsubscript{$\phi$\textsubscript{1}}(S)) > Q\textsubscript{$\theta$\textsubscript{1}}(S,\pi\textsubscript{$\phi$\textsubscript{1}}(S))
    \end{aligned}
\end{equation}
In such a case, the error increases as the first critic has a tendency to overestimate. Hence, to tackle this, a minimum function is used to avoid exaggeration. This may induce a component of underestimation bias due to underestimated action values, but this is preferred over overestimation.

\section{Results}
\label{15}
\begin{figure*}
    \includegraphics[width=\linewidth]{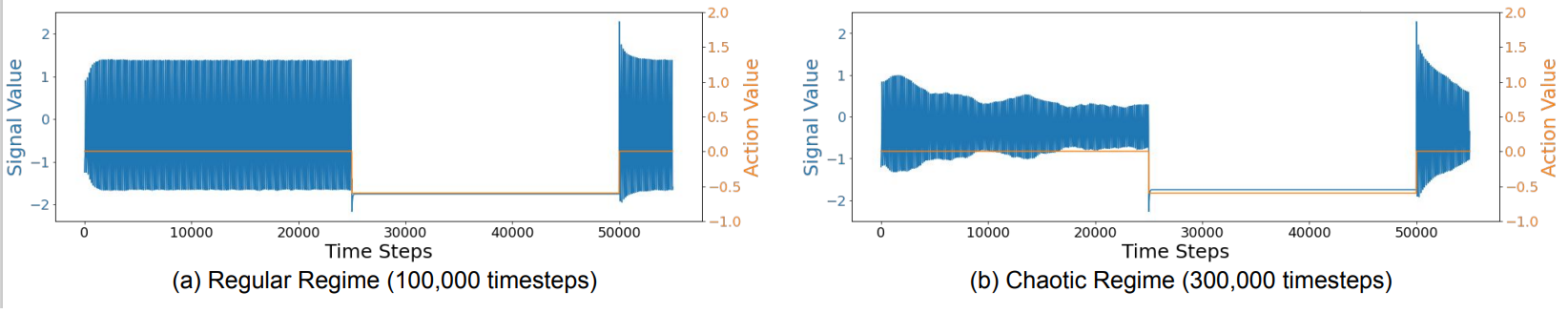}
\caption{Synchrony suppression in an ensemble 1000 neurons for regular and chaotic regime. The mean field (blue curve) and corresponding action pulses (orange curve) used for suppression are plotted (both curves have the same units but different corresponding scales).} \label{fig:TD3onenvs}
\end{figure*}

\begin{figure*}
    \includegraphics[width=\linewidth]{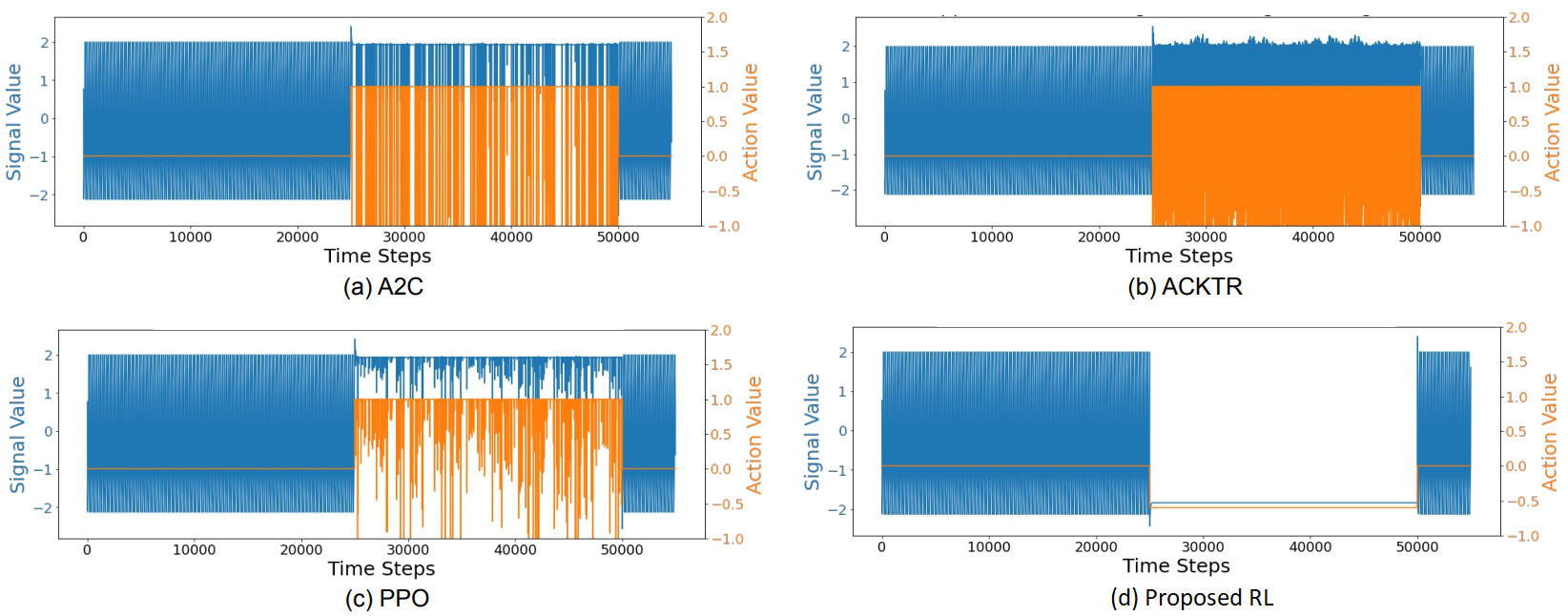}
\caption{Synchrony suppression in an ensemble 1000 neurons described by Hindmarsh-Rose model (bursting regime with $\epsilon$ = 0.2) for different models (500,000 steps). The mean field (blue curve) and corresponding action pulses (orange curve) used for suppression are plotted. (Both curves have the same units but different corresponding scales.)}                \label{fig:otheragentsonbursting}
\end{figure*}

The proposed agent was tested on an ensemble of 1000 regular (Bonhoeffer-van der Pol (Fig.~\ref{fig:TD3onenvs} (a)) neurons oscillating about an equilibrium point of $M\sim-0.2568$ with a coupling factor ($\epsilon$) of 0.03. This coupling factor decides the amplitude of the collective oscillation of neurons. The proposed agent is trained for 100,000 timesteps on the Regular regime. We commence synchrony suppression at $t$ = 250,000 time-steps by transmitting action pulses in agreement with our trained model. When the synchrony suppression is started, the oscillations spike a little in both negative and positive amplitude. After about 500 timesteps, steady state was reached and it converges around $M\sim-1.005$. Here, $M$ is defined as the Mean Point of Convergence which is essentially the mean of oscillations occurring at the steady state achieved by the model. The extent of suppression is determined by employing a suppression coefficient(s) based on the variance in the amplitude of oscillation~\cite{krylov2020}. 
\begin{equation}
    \begin{aligned}
        S = \frac{\sigma[M\textsubscript{before}]}{\sigma[M\textsubscript{after}]}
    \end{aligned}
\end{equation}
where $M\textsubscript{before}$ and $M\textsubscript{after}$ represent the mean of the field values before and after action pulse application. The energy required by the agent was calculated as the sum of modulus of actions taken by a particular agent. A high suppression coefficient of \num[round-precision=4,round-mode=figures, scientific-notation=true]{2489633367582914.5} was achieved.\\
\indent Next, the agent was tested on an ensemble of 1000 chaotic (Bonhoeffer-van der Pol (Fig.~\ref{fig:TD3onenvs}(b)) neurons and another ensemble of 1000 bursting (Hindmarsh-Rose (Fig.~\ref{fig:otheragentsonbursting}(d)) neurons. The neurons in the chaotic regime(coupling factor($\epsilon$) = 0.02) oscillate about a point $M\sim-0.2636$ whereas those in bursting regime(coupling factor($\epsilon$) = 0.2) oscillate about $M \sim-0.2772$. The proposed agent was trained for 300,000 timesteps on chaotic regime and for 500,000 timesteps on bursting regime. Again, synchrony suppression is started at $t = 250,000$ time-steps in both regimes, and similar spiking behaviour is observed for both regimes that settles about 500 time-steps after the commencement of suppression. 
It was observed that the convergence occurred around $M\sim-1.0054$ in case of the chaotic regime and around $M\sim-1.0563$ for the bursting regime. High suppression coefficients like \num[round-precision=4,round-mode=figures, scientific-notation=true]{3209418.7177} and \num[round-precision=4,round-mode=figures, scientific-notation=true]{6611661035639692.0} were achieved for chaotic and bursting regimes respectively. Suppression coefficient in bursting regime was incredibly suppressed compared to other RL algorithms such as Advantage actor critic (A2C), Actor critic with Kronecker-featured trust region (ACKTR) and proximal policy optimization (PPO) (See Figure \ref{fig:otheragentsonbursting}). Accordingly, the energy supplied (sum of the amplitude of action pulses) to the bursting ensemble was 49.47\% less compared to other RL algorithms (See Table 1). 


\begin{figure*}
    \includegraphics[width=\linewidth]{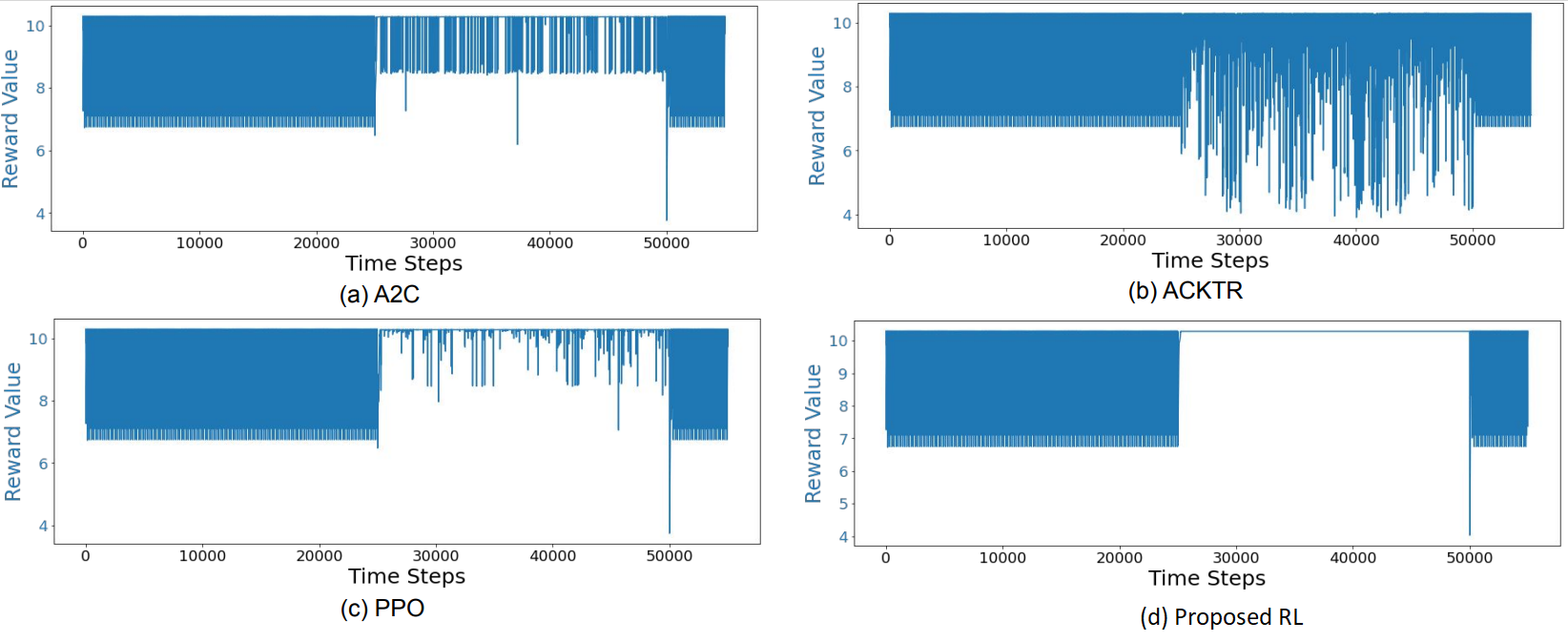}
\caption{Evaluation Rewards v/s Timesteps graph for various RL algorithms for bursting regime. The graph shows a high and stable reward when the model is applied to the environment, in contrast to oscillating rewards in the absence of the model.} 
    \label{fig:rewards}
\end{figure*}

Figure~\ref{fig:rewards} shows the rate of change evaluation rewards for various RL algorithms for bursting regime. The graph shows a high and stable reward when the proposed model is applied to the environment, in contrast to oscillating rewards in the absence of the model.

\begin{table}[]
 \caption{Performance metrics for all trained RL agents in Bursting Regime}
 \centering
\begin{tabular}{|c|c|c|}

\hline 
Model       & Energy Supplied to ensemble & M       \\ \hline \hline
A2C         & 24,991.03                   & 0.8141  \\ \hline
ACKTR       & 20,476.76                   & 0.6326  \\ \hline
PPO         & 24,855.793                  & 0.8201  \\ \hline
\textbf{Proposed RL} & \textbf{14,996.508}         & \textbf{-1.0563} \\ \hline
\end{tabular}
\end{table}
\section{Conclusion}
\label{co}
In this paper, a novel RL algorithm is proposed for suppressing the synchronisation for neurological patients using closed loop DBS. The proposed model was able to suppress neural synchrony using an ensemble of Bonhoeffer-van der Pol oscillators and Hindmarsh-Rose oscillators. The proposed work showed better suppression capability (a higher suppression coefficient) with fewer time stamps as compared to A2C, ACKTR, and PPO, but at the cost of elevated energy consumption. For future scope, this work will be updated to implement variable reward schedules, such as variable ratios, fixed schedules, and variable schedules. Additionally, the effect of these schedules compared to a fixed ratio will be studied. Also, in the future, we plan to test the algorithm on a "brain-on-chip" model so as to qualitatively study its performance on actual hardware. RL, being a complex model, requires more power. In the future, we will attempt to reduce this power consumption without compromising the suppression coefficient.


\end{document}